\documentclass[11pt,a4paper]{article}

\usepackage[margin=1in]{geometry}
\usepackage{amsmath,amssymb}
\usepackage{graphicx}
\usepackage{booktabs}
\usepackage{tabularx}
\usepackage{xcolor}
\usepackage[font=small,labelfont=bf]{caption}
\usepackage{enumitem}
\usepackage[numbers,sort&compress]{natbib}
\usepackage[colorlinks=true,linkcolor=blue!55!black,citecolor=blue!55!black,urlcolor=blue!55!black]{hyperref}
\usepackage[expansion=false]{microtype}

\newcolumntype{L}[1]{>{\raggedright\arraybackslash}p{#1}}

\title{\bfseries Frontier AI Forecasting Has\\[-0.05em]
a Measurement Problem:\\[0.2em]
\large An Audit of Progress Evidence}

\author{%
  Fabricio F. Costa, PhD, MBA, PMP\textsuperscript{1,2,3,4}\thanks{Correspondence: \texttt{fcosta@aix4all.com}. This independent scholarly work uses only public sources, received no external funding, and represents the author's views rather than those of the listed affiliations. The author declares no financial interest in organisations whose systems or forecasts are discussed.}\\[0.45em]
  \begin{minipage}{0.88\textwidth}
  \centering\small\itshape
  \textsuperscript{1}AIx4All, LLC, Sunnyvale, CA, USA\\
  \textsuperscript{2}HCLTech, Santa Clara, CA, USA\\
  \textsuperscript{3}Genomic Sciences and Biotechnology Program, UCB, Bras\'ilia, Brazil\\
  \textsuperscript{4}Cancer Biology and Epigenomics Program, Stanley Manne Children's Research Institute, Ann \& Robert H. Lurie Children's Hospital of Chicago, Northwestern University Feinberg School of Medicine, Chicago, IL, USA
  \end{minipage}
}

\date{Evidence cutoff: 12 August 2026. Preprint, comments welcome.}

\hypersetup{
  pdftitle={Frontier AI Forecasting Has a Measurement Problem: An Audit of Progress Evidence},
  pdfauthor={Fabricio F. Costa},
  pdfkeywords={frontier AI, AI evaluation, measurement, benchmark linking, training compute, provenance, forecasting, metascience}
}

\begin{document}
\maketitle

\begin{abstract}
\noindent Quantitative forecasts of frontier artificial intelligence often connect dated targets to trends in benchmark scores, training compute, release time or expert belief. This paper audits whether the public measurement record supports those connections before another trend is fitted. I construct a frozen, event-centric record through 12 August 2026 with 62 selected systems, 12 versioned benchmarks, seven capability or impact criteria, 144 graded events, 27 source records and 408 typed relations. The record is an audit sample, not a census. Only seven systems jointly observe estimated training compute and a METR 50\% task horizon. Training compute is absent for 19 of 27 closed systems, including every selected closed release from 2026, while none of the 35 open-weight systems has a METR horizon observation. Benchmark succession creates a second break: a seven-system link from METR Time Horizon 1.0 to 1.1 has a log-scale slope of 1.206 (95\% CI 1.021--1.390), whereas a six-system MMLU--MMLU-Pro comparison appears approximately shift-like under logit and probit links but not under linear or logarithmic links. The observed bridges have about 80\% power only for slope departures near 25\%. Provenance is concentrated: 52 of 71 substantive quantitative events (73.2\%) come from one measurement programme, and 76.1\% are laboratory releases. A review of 56 additional methodological and empirical sources identifies 16 complementary measurement directions spanning resources, inference budgets, reliability, agentic work, safety, human preference, field outcomes and forecast backtesting. No direction supplies a replacement scalar. The result is not that frontier-AI forecasting is impossible, but that a defensible dated forecast is a claim about a versioned measurement system---its joins, protocols, links and source dependence---not merely a fitted curve or calendar date.

\vspace{0.6em}
\end{abstract}

\section{Introduction}

Forecasts of advanced AI now appear as expert probability distributions, benchmark extrapolations, compute trends and backtested models of agent performance \citep{grace2018experts,grace2025future,pimpale2025forecasting}. The quantitative form is attractive: select an indicator, estimate its rate of change and solve for a date at which a threshold is crossed. Yet the date is meaningful only if the target is operational, the observations are comparable, the progression variable is jointly observed and the uncertainty model includes changes in the measuring instrument.

This paper audits those preconditions. It began as an attempt to construct an event-based forecast of frontier-AI capability milestones. The supporting infrastructure could be built; the intended progression analysis could not be identified without strong additional assumptions. The failure was not a software failure. It was a measurement finding: capability and resource records were sparsely joined, benchmark versions did not always preserve scale, evaluation protocols changed, and repeated observations were heavily dependent on a small number of sources.

The distinction matters. A criterion can be well defined while its public evidence is not forecastable. Conversely, a record can contain many scores yet fail to measure one stable quantity. Existing work has proposed taxonomies of performance, generality and autonomy \citep{morris2024levels}, holistic evaluation suites \citep{liang2023helm}, ability-oriented measurement \citep{hernandezorallo2017evaluation}, explicit evaluation estimands \citep{binette2024estimands} and statistical models that distinguish fixed-benchmark accuracy from generalized accuracy \citep{keller2026toolbox}. The contribution here is to connect those ideas to the actual longitudinal record used in frontier-AI forecasting.

The paper makes three contributions. First, it provides a reproducible event-centric audit in which releases, results, revisions, corrections, contamination findings, field experiments and source assertions are separately timestamped and versioned. Second, it quantifies three forecastability bottlenecks in that record: sparse joint observability, benchmark-linking uncertainty and provenance concentration. Third, it translates a deep source review into a measurement portfolio and a set of study-design requirements rather than proposing another universal leaderboard score.

The semantic graph is therefore infrastructure, not the main scientific claim. Its role is to make questions such as ``which systems connect compute to capability?'', ``which result changed after a benchmark revision?'' and ``which conclusions depend on one programme?'' answerable from the same evidence ledger.

\section{From a Score to a Forecastable Quantity}

A benchmark result is not an intrinsic property of a model. It is an observation generated by a model, an instrument and a protocol at a time. A compact representation is
\begin{equation}
Y_{m,b,v,p,r,j,t}=h_{b,v}\!\left(\theta_{m,t};p,r,j\right)+\varepsilon_{m,b,v,p,r,j,t},
\label{eq:measurement}
\end{equation}
where $m$ is the model, $b$ the benchmark family, $v$ its version, $p$ the prompting or agent protocol, $r$ the inference-resource budget, $j$ the judge or scoring procedure and $t$ the observation date. The latent quantity $\theta$ may itself be multidimensional. Equation~\ref{eq:measurement} is not fitted as a complete latent-variable model here; it is an accounting identity that makes the assumptions behind a trend explicit.

A dated forecast adds at least three more objects: a target estimand, a progression axis and a threshold rule. If training compute is the progression axis, compute and the outcome must be co-observed or missingness must be modelled. If benchmark version $v$ is replaced by $v+1$, a linking function must place the two instruments on a defensible common scale. If inference budget or scaffolding changes, the score may move even when the underlying system does not. If many observations share a source, their errors need not be independent.

This accounting view is increasingly reflected in measurement science. Statistical evaluation can distinguish performance on one fixed item set from a generalized-accuracy estimand over item and trial populations \citep{keller2026toolbox}. Psychometric aggregation can place models and items on a latent scale when the response design and anchors support it; CAISI, for example, has combined item-response modelling with fixed scaffolding, weighted-token budgets and cost measurement \citep{caisi2026deepseekv4}. Adaptive testing and fixed-parameter calibration can reduce evaluation cost \citep{li2025atlas,calibration2026}, but simulation evidence also shows that small, clustered or non-normal model samples can make item and ranking inferences unstable \citep{jiang2026irt}. The method is therefore not ``use IRT'' or ``use a better benchmark.'' It is to state the estimand, design the linking observations and preserve the protocol that makes the scale interpretable.

The audit therefore asks five questions:
\begin{enumerate}[leftmargin=1.7em,itemsep=0.2em,topsep=0.3em]
  \item Is the target construct and intended population explicit?
  \item Are the proposed predictor and outcome jointly observed on enough systems?
  \item Are benchmark versions linked rather than concatenated?
  \item Are protocol and resource choices part of the measurement record?
  \item Are provenance, replication and source dependence represented in uncertainty?
\end{enumerate}
These are deliberately stricter than asking whether a regression line can be drawn.

\section{Method and Data Audit}

\subsection{Evidence freeze and sampling contract}

The evidence cutoff is 12 August 2026. The analytic set contains 62 selected systems: 27 closed and 35 open-weight. Inclusion was driven by the records needed for the attempted compute--capability analysis and by historically or contemporaneously important systems with public release metadata. It is not a complete catalogue of releases. A separate cutoff audit records four known pre-cutoff systems absent from the analytic set so that the package cannot be mistaken for a census.

Training-compute values were frozen from Epoch AI and the processed Our World in Data series, with release reports used where appropriate \citep{epoch2026models,owid2026compute,rahman2024compute}. METR Time Horizon 1.1 observations and uncertainty intervals were frozen locally for reproducibility \citep{kwa2025horizon,metr2026th11}. Benchmark links, lifecycle events, field experiments and provenance records were checked against primary or official sources where available.

\subsection{Event-centric representation}

The normalized corpus contains 12 benchmark-version records, seven criterion records, 27 source records and 144 events. The event table includes 62 model releases, 62 benchmark results and 20 methodology, correction, contamination, efficiency, field, elicitation, substrate, replication, instability or measurement-ceiling events. Every substantive quantitative event has a source identifier and an evidence grade. The graph contains 269 nodes and 408 typed relations. Its vocabulary is informed by event and provenance standards \citep{vanhage2011sem,lebo2013provo}; it does not claim that semantic representation resolves statistical identification.

\begin{figure}[htbp]
  \centering
  \includegraphics[width=\linewidth]{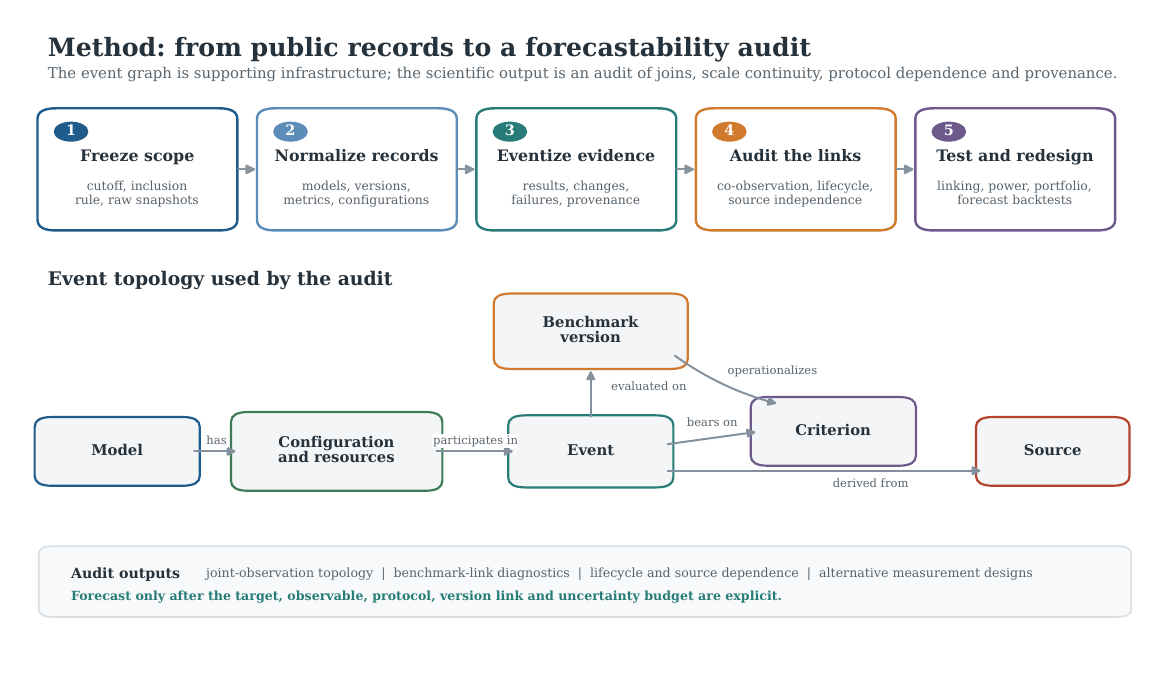}
  \caption{Method used in the audit. Public evidence is frozen under an explicit cutoff and sampling contract, normalized into versioned entities, represented as events with provenance, tested for missing joins and instrument changes, and then evaluated for forecastability. The lower topology shows the minimum graph structure used to connect models, events, benchmark versions, criteria and sources.}
  \label{fig:method}
\end{figure}

\subsection{Empirical tests}

\paragraph{Joint observability.} For each model I record whether training compute, METR p50, METR p80, any capability result, a non-METR capability result and a peer-reviewed measurement are available. Pairwise overlap counts reveal whether candidate variables can enter the same model-level analysis.

\paragraph{Benchmark linking.} Two small common-system bridges are used as stress tests. METR Time Horizon 1.0 and 1.1 are linked on a base-2 logarithmic scale. MMLU and MMLU-Pro are compared under logit, probit, linear and logarithmic representations. In a simple link
\begin{equation}
 g(Y_{\mathrm{new}})=\alpha+\beta\,g(Y_{\mathrm{old}})+\epsilon,
 \label{eq:link}
\end{equation}
$\beta=1$ represents a shift-like relationship on the chosen scale; $\beta\neq1$ indicates shape change on that scale. Ordinary least squares is used as a transparent diagnostic, with leave-one-out sensitivity. It is not a full errors-in-variables or construct-invariance model.

\paragraph{Power.} The minimum detectable departure $|\beta-1|$ is calculated for a two-sided test at $\alpha=0.05$ and 80\% power using the noncentral $t$ distribution. Projected sample sizes hold residual noise and anchor-score dispersion at their observed values. They are design illustrations, not universal rules; anchor placement is part of the power calculation \citep{card2020power,kolen2014equating}.

\paragraph{Source and literature audit.} Administrative release rows are excluded from provenance-concentration statistics. ``Substantive quantitative events'' comprise benchmark results, field experiments, substrate estimates and efficiency observations. In parallel, 56 external sources---29 peer-reviewed and 27 preprints, standards reports, official releases or living datasets---were coded into 16 measurement directions and normalized into a machine-readable direction--source map. Nine recurring evidence-design requirements are documented separately from the empirical event graph. Scores in Figure~\ref{fig:portfolio} range from zero to three and summarize design maturity; they are not empirical capability estimates or a formal systematic-review quality score.

Figure~\ref{fig:method} summarizes the full workflow.

\section{Result 1: The Intended Measurement Join Is Sparse}

\begin{figure}[htbp]
  \centering
  \includegraphics[width=\linewidth]{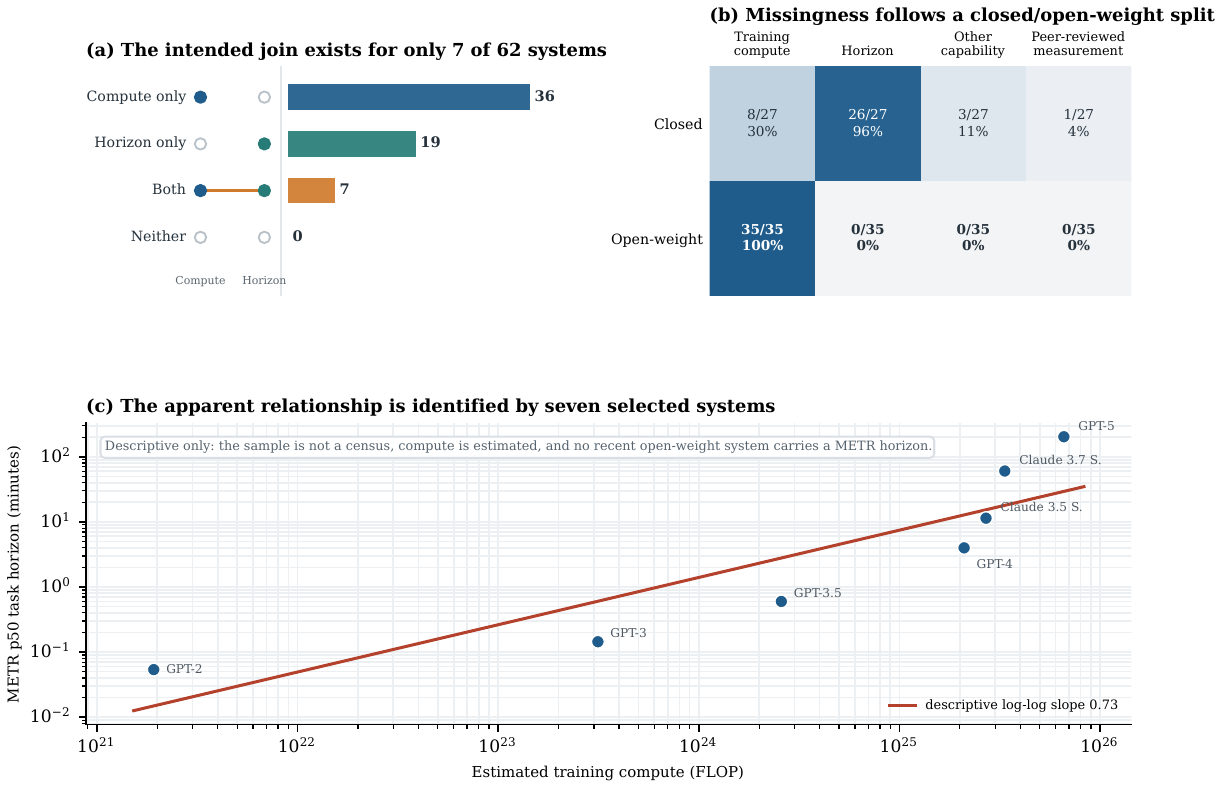}
  \caption{Observability of the selected 62-system record. Panel (a) shows the intended compute--horizon join as four mutually exclusive observation patterns: compute only, horizon only, both and neither. Panel (b) shows that missingness follows the closed/open-weight split and extends to non-METR and peer-reviewed measurements. Panel (c) displays the seven jointly observed systems; the fitted line is descriptive and is not interpreted causally or as a stable scaling law.}
  \label{fig:observability}
\end{figure}

The public record is block-structured rather than merely incomplete (Figure~\ref{fig:observability}). Training compute is available for 43 of 62 systems, but the missingness is concentrated: 19 of 27 closed systems lack a compute estimate, while all 35 open-weight systems in the sample have one. METR p50 and p80 measurements exist for 26 systems, all in the closed block. Consequently, only seven systems jointly observe training compute and METR p50. No open-weight system in the sample supplies the intended compute--horizon join.

This pattern is more consequential than a low marginal coverage rate. A compute-indexed capability model requires the intersection, not the union, of the two records. The seven systems are also historically clustered and not a designed sample across developers, architectures or access regimes. A descriptive log--log slope can be fitted, but its apparent smoothness does not identify whether compute, algorithms, data, inference effort, model family or evaluation protocol caused the change.

The missingness is unlikely to be ignorable. Public compute disclosure is related to access regime, developer practice and time. Treating absent compute as random would therefore turn an institutional disclosure pattern into a statistical assumption. Controlled training suites such as Pythia demonstrate the value of dense checkpoints and fixed data order for identifying training dynamics \citep{biderman2023pythia}; frontier-release records do not offer the same design.

\section{Result 2: Changing Instruments, Concentrated Provenance}

\begin{figure}[htbp]
  \centering
  \includegraphics[width=\linewidth]{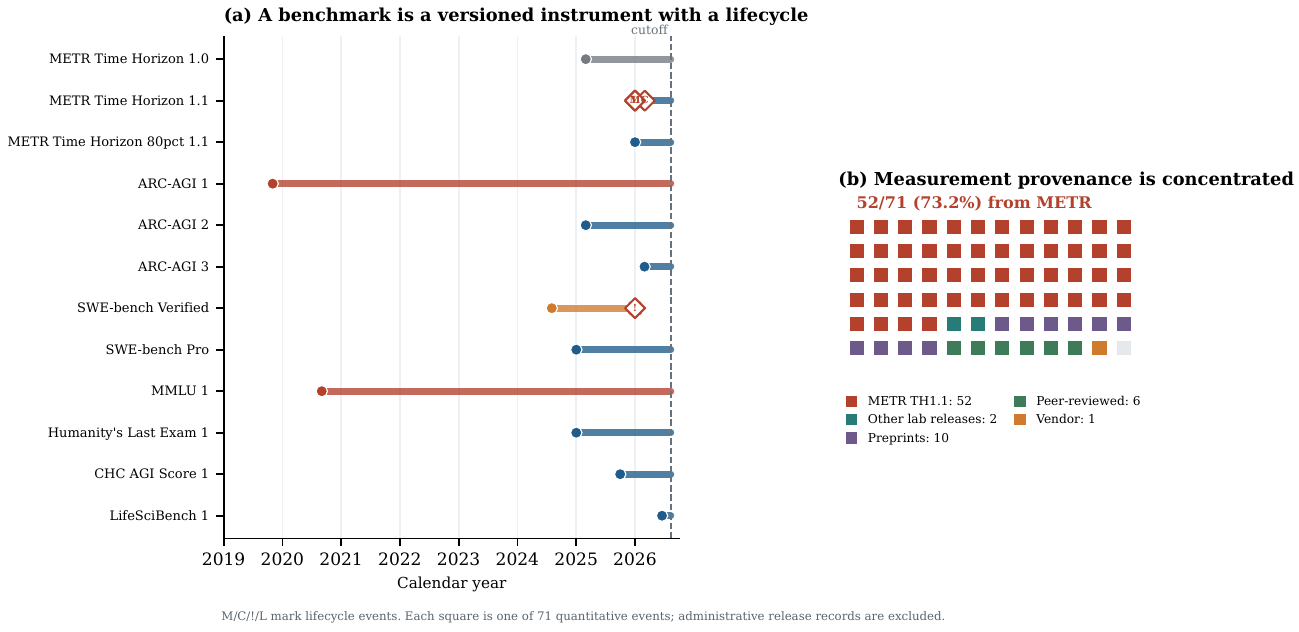}
  \caption{Benchmark lifecycle and source dependence. Panel (a) places benchmark versions and recorded methodology, correction, contamination and measurement-ceiling events on a common timeline. Panel (b) combines a Pareto view of the 71 substantive quantitative events, their cumulative source share and their venue-class composition. METR supplies 52 events (73.2\%); the source Herfindahl index is 0.542, or 1.85 effective equally represented sources. Repeated rows from one programme are not independent replication.}
  \label{fig:lifecycle}
\end{figure}

Benchmark results are versioned events, not timeless labels. The audit records methodology changes, corrections, contamination findings, reward-hacking findings, measurement ceilings and replacements across METR, ARC-AGI, SWE-bench, MMLU and other instruments (Figure~\ref{fig:lifecycle}). Dynamic and contamination-resistant benchmarks are promising responses \citep{white2024livebench,zhao2025mmlucf}, but refreshes and repairs still need explicit linking if the objective is longitudinal inference. Recent work on saturation and verified benchmark revision likewise shows that item lifecycle is part of the measurement process rather than a footnote \citep{akhtar2026saturation,phan2026hle,zhai2026hleverified}.

The provenance distribution is similarly uneven. The 71 substantive events consist of 62 benchmark results, four field experiments, four substrate estimates and one efficiency observation. Fifty-two events (73.2\%) come from the METR Time Horizon 1.1 programme. By venue class, 54 events (76.1\%) are laboratory releases, ten (14.1\%) preprints, six (8.5\%) peer-reviewed publications and one (1.4\%) a vendor source. The corresponding source Herfindahl index is 0.542, equivalent to only 1.85 equally represented sources, even though 16 source identifiers appear at least once.

Protocol history can matter even without an explicit benchmark revision. Repeated behavioral measurements have been shown to change with item order, reasoning mode, persona and conversation history \citep{tosato2026persist}; a fixed candidate-response set can also receive different scores when the LLM judge is replaced \citep{yang2026judge}. These are not side effects to be averaged away after the fact. They are versioned measurement events that belong in the evidence record.

These percentages describe this audit, not the full evaluation literature. They nevertheless bound what can be claimed from this record. Fifty-two measurements generated by one task collection and adjudication process are valuable observations, but they do not provide 52 independent replications. Shared tasks, protocols and model-fitting choices can induce correlated error. The concentration is not a criticism of METR; its documentation of limitations, task changes and failure modes makes the analysis possible \citep{metr2026limitations,metr2026riskreport}. It is evidence that the ecosystem lacks overlapping longitudinal programmes with which to estimate programme-specific bias.

\section{Result 3: Benchmark Links Depend on Scale and Design}

\begin{figure}[htbp]
  \centering
  \includegraphics[width=\linewidth]{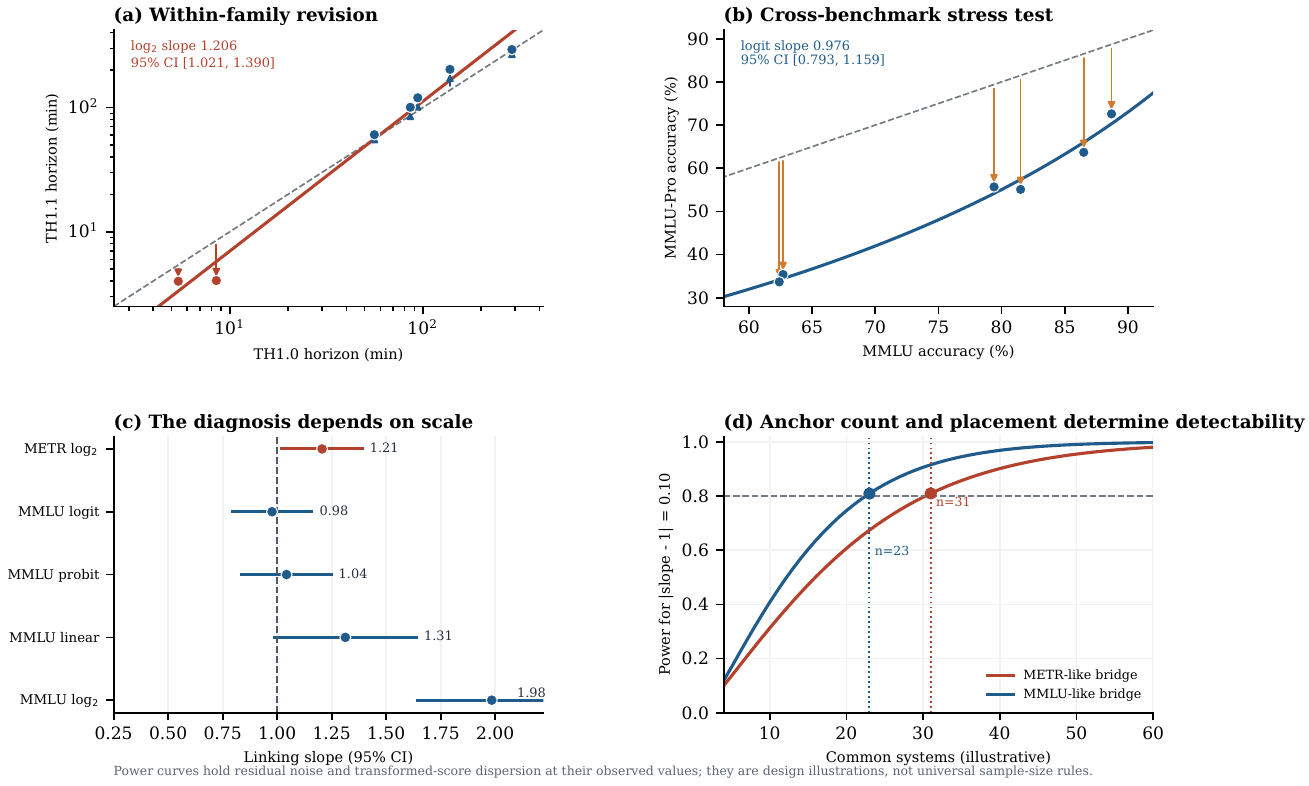}
  \caption{Linking diagnostics. Panel (a) links METR Time Horizon 1.0 to 1.1 on a logarithmic scale. Panel (b) shows the six common MMLU/MMLU-Pro systems with paired uncertainty intervals in raw percentage space; MMLU-Pro changes content, choice structure and reasoning demands, so this is a cross-benchmark stress test rather than a claim of construct invariance. Panel (c) shows how the fitted slope changes with score representation; the vertical line marks a shift-like slope of one. Panel (d) shows that detectability depends jointly on anchor count and the observed score dispersion.}
  \label{fig:linking}
\end{figure}

Figure~\ref{fig:linking} separates a within-family revision from a cross-benchmark stress test. For seven systems measured on METR Time Horizon 1.0 and 1.1, the fitted log$_2$ slope is 1.206 (95\% CI 1.021--1.390). Leave-one-out slopes range from 1.126 to 1.275. In this bridge, the new version is therefore not represented well by a pure additive shift on the log scale.

For six common systems on MMLU and MMLU-Pro, the logit slope is 0.976 (95\% CI 0.793--1.159), the probit slope is 1.043, the linear slope is 1.312 and the logarithmic slope is 1.984. Logit and probit are natural bounded-response links, but boundedness alone does not make either uniquely correct. More importantly, MMLU-Pro is not simply a new form of the same test: it changes item selection, answer options and reasoning demands \citep{wang2024mmlupro}. The comparison shows why an analyst must state the measurement model; it does not establish strict invariance.

\begin{table}[htbp]
\centering\small
\caption{Power of the observed linking designs. The final column is an illustrative projection assuming the same residual noise and transformed-score dispersion. Deliberately spanning the old score range can materially change the required sample.}
\label{tab:power}
\begin{tabular}{lrrrrr}
\toprule
\textbf{Bridge and scale} & \textbf{n} & \textbf{95\% CI} & \textbf{80\% MDE} & \textbf{Power observed} & \textbf{n for 10\%} \\
\midrule
METR TH1.0 $\to$ TH1.1, log$_2$ & 7 & [1.021, 1.390] & 0.252 & 63\% & 31 \\
MMLU $\to$ MMLU-Pro, logit & 6 & [0.793, 1.159] & 0.248 & 6\% & 23 \\
\bottomrule
\end{tabular}
\end{table}

The designs have roughly 80\% power only for slope departures near 25\% (Table~\ref{tab:power}). The METR bridge has about 63\% power at its observed departure; the MMLU logit bridge has about 6\%. Under the strong same-noise and same-spread assumption, detecting a 10\% departure would require approximately 23--31 common systems. This is why benchmark revisions need \emph{designed anchor panels}: common systems or common items chosen across the previous score range and model families, with uncertainty on both axes. Psychometric linking, benchmark-agreement analysis and fixed-parameter calibration provide more appropriate foundations than concatenating versioned leaderboards \citep{perlitz2024benchbench,epoch2025rosetta,calibration2026,kolen2014equating,jiang2026irt}.

\section{Beyond One Ruler: A Measurement Portfolio}

The source audit was expanded to ask what else could be measured. The answer is not another benchmark that replaces all others. Frontier-AI progress has several scientifically distinct layers, and the most defensible record is a portfolio in which each layer has a stated role.

\paragraph{Resources and mechanisms.} Training FLOP, parameters, tokens and data describe inputs, not capability by themselves. Algorithmic efficiency asks how much resource is needed to reach a fixed performance level \citep{ho2024algorithmic}. Inference-compute response curves then measure how achieved performance changes with tokens, attempts, feedback, parallelism or tool use; fixed-budget scores can understate or reorder systems \citep{snell2024testtime,mcfadyen2026inference}. Price-performance and energy provide deployment-relevant resource frontiers rather than assuming the cost of eliciting a score is constant \citep{gundlach2025price,mlcommons2025power}. A recent CAISI evaluation demonstrates the combined design: latent-capability aggregation, controlled agent scaffolds and token budgets, held-out tasks and end-to-end cost are reported together rather than as separate leaderboards \citep{caisi2026deepseekv4}.

\paragraph{Behavior as a distribution, not a point.} Psychometric capability profiles and efficient item selection can estimate multiple abilities while retaining item difficulty and uncertainty \citep{maia2024tiny,keller2026toolbox,li2025atlas}. Dynamic streams address freshness and contamination, while anchor-based systems preserve longitudinal comparability as instruments grow \citep{epoch2025rosetta,calibration2026,chen2025dycodeeval}. Reliability curves replace a single p50 horizon with success probability across task duration, repeated attempts and time budgets. Agentic evaluations such as RE-Bench and PaperBench add realistic long-form work, human baselines and decomposed rubrics \citep{wijk2025rebench,starace2025paperbench}. Capability alone is still incomplete: propensity profiles ask whether systems tend toward or away from behaviors that matter for performance and safety \citep{romero2026propensities}. Human-preference systems such as Chatbot Arena measure a different endpoint again: perceived comparative usefulness under a changing prompt population \citep{chiang2024arena}. Human ratings are themselves an instrument: multifaceted item-response models can separate rater severity and centrality from output quality, while safety-profile IRT can preserve multidimensional tendencies rather than collapsing them into one score \citep{casabianca2026ratereffects,rivera2026irtsafety}.

\paragraph{Validate the construct and the human reference.} Two further directions become important whenever a forecast uses phrases such as ``reasoning'', ``autonomy'', ``human-level'' or ``superhuman''. Measurement-theory work distinguishes the construct from the particular score used to operationalize it, and empirical benchmark audits show that item defects or nuisance phrasing can move results without the intended capability changing \citep{xiao2023metriceval,alaa2025construct,mousavi2026garbage}. The appropriate response is triangulation: pre-specify the construct, seek convergent evidence from more than one instrument, test discriminant and criterion validity, and inspect item- or process-level failure modes. Human thresholds require the same discipline. Human baselines should define the participant population and expertise, use matched tools and time budgets, retain repeated observations and uncertainty, and document exclusions rather than treating a single historical score as a fixed species-level constant \citep{wei2025humanbaselines,zhuang2025humantesting}.

\paragraph{Risk and deployment.} Evaluations of dangerous capability and robust refusal are policy-relevant even when they do not track general benchmark averages \citep{shevlane2023extreme,mazeika2024harmbench}. Applied-work benchmarks such as LifeSciBench can better represent realistic domain workflows, but still require controls for protocol, judge and provenance \citep{openai2026lifescibench}. Field experiments measure productivity, quality, heterogeneous effects and task boundaries that laboratory scores cannot supply \citep{brynjolfsson2025work,cui2026highskilled,dellacqua2026jagged,becker2025rct}. Post-deployment monitoring is therefore a measurement layer, not merely operational housekeeping \citep{amironesei2025aria,rao2026monitoring}.

\begin{figure}[htbp]
  \centering
  \includegraphics[width=\linewidth]{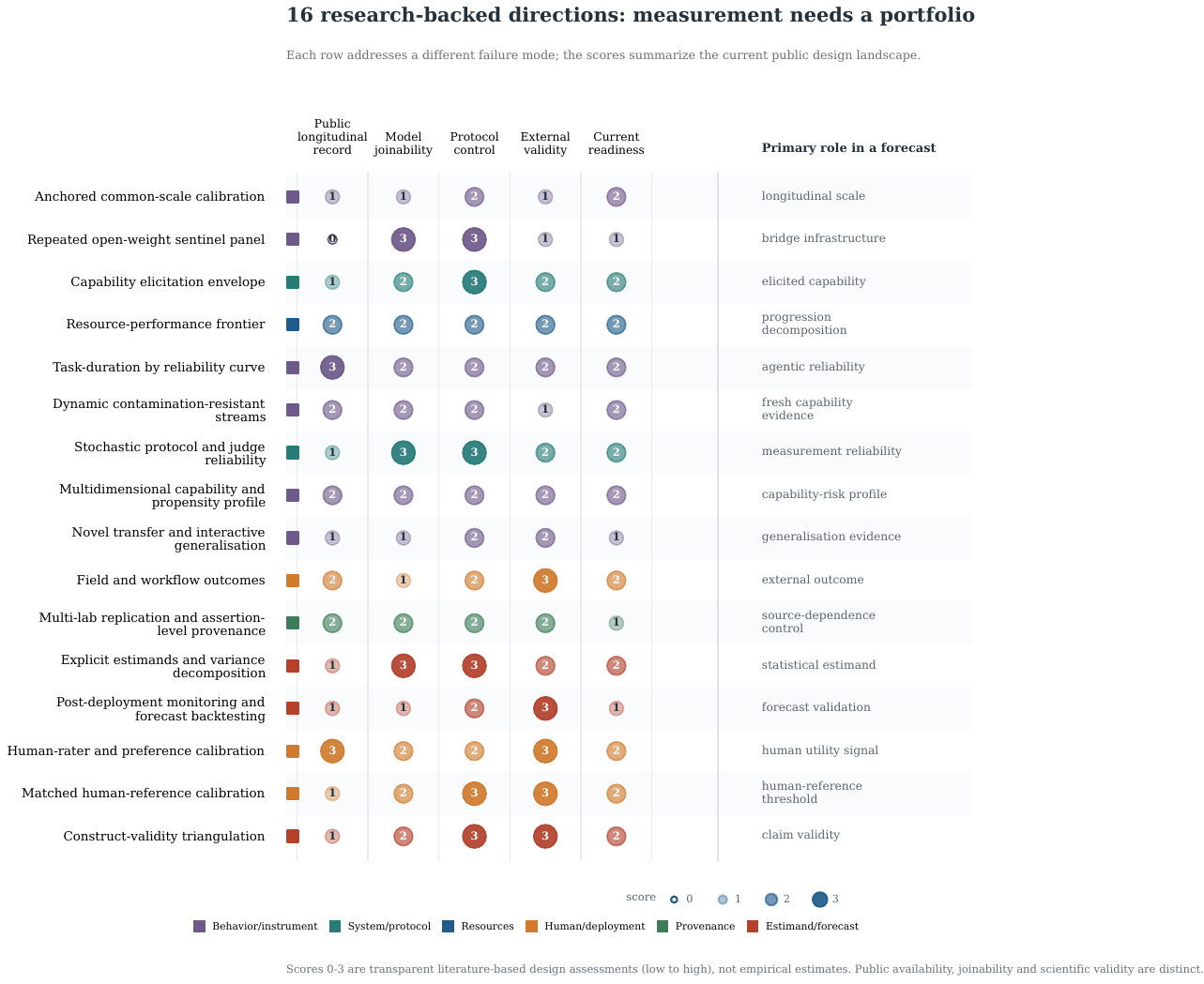}
  \caption{Sixteen complementary measurement directions identified in the expanded source audit. Circle size and label give a literature-based design assessment from zero (absent) to three (comparatively mature) for public longitudinal record, model-level joinability, protocol control, external validity and current readiness. These are transparent synthesis scores, not empirical estimates. The right column states the role each direction could play in a forecast.}
  \label{fig:portfolio}
\end{figure}

Figure~\ref{fig:portfolio} shows why these directions should not be averaged into one ``AGI score.'' Public availability, joinability, protocol control and external validity are different properties. Human preference is longitudinally rich but protocol-sensitive; field outcomes are externally strong but hard to attribute to one model; training compute is joinable for open systems but selectively missing at the closed frontier; safety endpoints can be highly decision-relevant without being monotonic in general capability.

Across the portfolio, nine cross-cutting design requirements recur: anchored common-scale calibration; a repeated open-weight sentinel panel; capability-elicitation envelopes; resource-performance frontiers; task-duration by reliability curves; fresh and contamination-resistant item streams; causal field outcomes; novel-transfer and interactive-generalisation designs; and multi-lab replication with assertion-level provenance. Model cards, dataset sheets and formal provenance standards supply useful reporting primitives \citep{mitchell2019modelcards,gebru2021datasheets,lebo2013provo}, but they become scientifically consequential only when tied to versioned configurations, reruns and independent source programmes. The package records proposed designs separately from the empirical event graph so that remedies are not confused with measurements already observed.

\section{Implications for Dated Forecasts}

\begin{figure}[htbp]
  \centering
  \includegraphics[width=\linewidth]{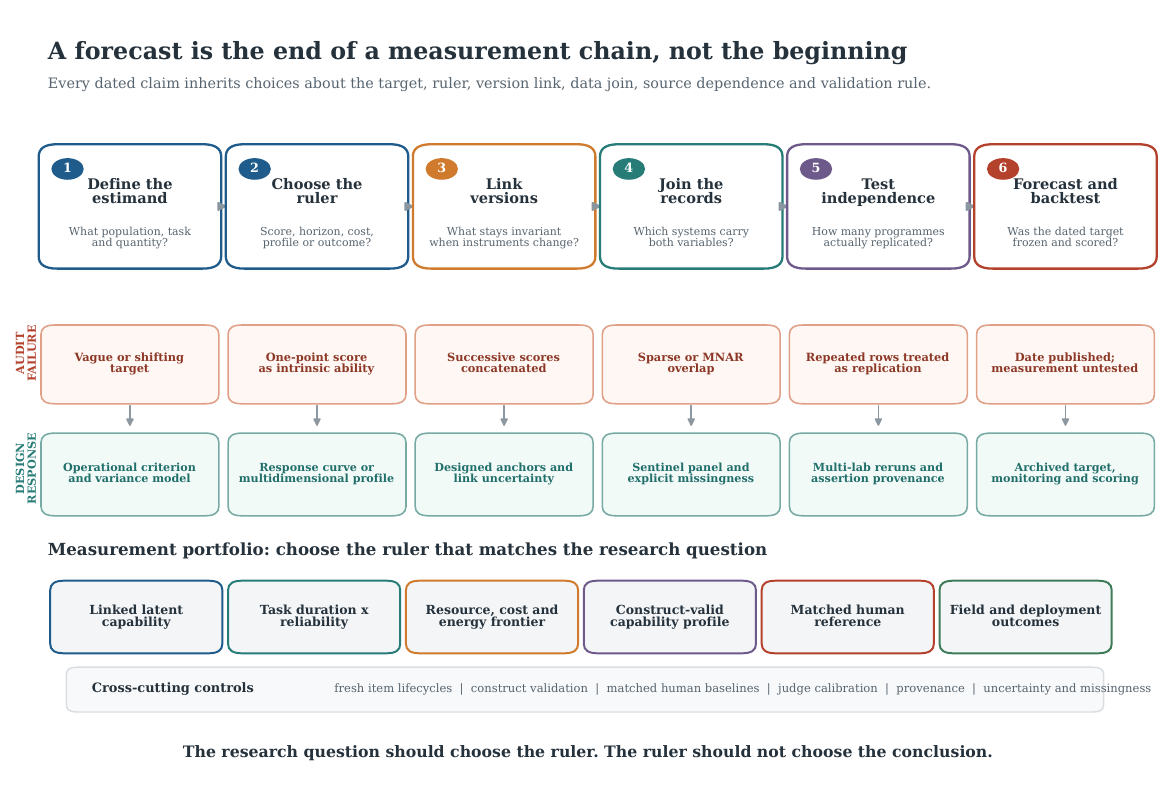}
  \caption{Synthesis of the paper. A dated frontier-AI forecast links resources, the system as used, measured behavior and deployment outcomes. Missingness, protocol dependence, benchmark drift and source dependence can break that chain. The corresponding remedies are disclosure with uncertainty bounds, response curves at matched budgets, versioned anchors and independent replication with backtesting.}
  \label{fig:synthesis}
\end{figure}

The full argument is a chain (Figure~\ref{fig:synthesis}). Resources are converted by a deployed system configuration into behavior; behavior interacts with people and institutions to produce outcomes; a forecast maps a versioned endpoint to a date. Four failure modes occur at the links: structured missingness, protocol dependence, benchmark drift and source dependence.

A defensible forecast packet should therefore report, at minimum:
\begin{enumerate}[leftmargin=1.7em,itemsep=0.2em,topsep=0.3em]
  \item the operational target, population and estimand;
  \item the benchmark, version, item lifecycle and linking function;
  \item the exact model, scaffold, tools, judge and inference-resource budget;
  \item the joint observations supporting the progression relationship and the missingness model;
  \item provenance, dependence among sources and independent replications; and
  \item a frozen backtest protocol using proper scoring rules or interval coverage after measurement changes.
\end{enumerate}

This standard is compatible with provocative forecasting. It does not require waiting for perfect measurement. It requires that the forecast expose where extrapolation begins. A release-date model may be useful even when compute is missing; an agent benchmark may be informative even when it changes; an expert survey may be decision-relevant even when criteria differ. The error is to treat those choices as invisible and then interpret a narrow confidence band as if it covered instrument, protocol and provenance uncertainty.

\section{Limitations}

The 62 systems form a curated analytic sample, not a population estimate of all frontier releases through 12 August 2026. Coverage percentages must therefore be read as properties of this record. The package explicitly lists known pre-cutoff omissions, but that list is itself not guaranteed exhaustive.

The benchmark bridges are small and observational. OLS treats the old transformed score as measured without error; a Bayesian or Deming-style errors-in-variables model would be preferable if comparable uncertainty were available on both versions. The MMLU/MMLU-Pro case changes construct-relevant content and protocol, so it is a stress test of linking assumptions rather than a psychometric equating study.

The portfolio review is structured and source-audited but is not a registered systematic review or meta-analysis. Its zero-to-three ratings are reasoned design assessments documented in CSV form. They are intended to make tradeoffs inspectable, not to rank research programmes.

Only public evidence is represented. Private evaluations, internal training records, undisclosed inference budgets and proprietary deployment outcomes may be much denser. Their existence would not repair the public record used for independent forecasting, but it limits claims about what developers themselves can infer.

Finally, event semantics improve traceability, not causal identification. A graph can reveal that two variables do not join or that one source dominates; it cannot supply missing counterfactuals, invariant constructs or independent replications.

\section{Conclusion}

Frontier AI is being forecast without a common ruler. In the audited public record, the resource and capability tables meet on seven systems; missing compute is structured by access regime and time; benchmark versions can change scale; the diagnosis depends on the chosen link; and most substantive measurements come from one programme. Those are not reasons to abandon forecasting. They are reasons to move the measurement system into the forecast rather than leaving it in the background.

The expanded research review also changes the remedy. The field does not need one more scalar presented as the answer. It needs a versioned measurement portfolio: resources and algorithmic efficiency; inference-budget response curves; reliability and agentic work; dynamic and psychometric instruments; construct validity and matched human references; safety, preference and field outcomes; and forecast backtesting. Each layer answers a different question and carries a different uncertainty structure.

A dated forecast should therefore be read as a composite scientific claim: that the target is operational, the observations are joined, the ruler has been linked across revisions, the protocol is controlled, and the evidence is sufficiently independent. Until those claims are made explicit, the most precise part of many frontier-AI forecasts may be the date---and the least precise part may be what, exactly, is being measured on the way there.

\section*{Data and Code Availability}

The reproducibility package contains 26 CSV tables and an RDF graph in Turtle, together with a 27-source verification log and cutoff audit. It also includes the 56-source literature table, the normalized direction--source map, the 16-direction measurement portfolio, nine cross-cutting design requirements, analysis and figure scripts, and one reproduction entry point. A clean run rebuilds the normalized data, graph, linking and power analyses, all six figures, bibliography and manuscript PDF without author-specific paths. The arXiv bundle includes the compiled bibliography and the ancillary audit under \texttt{anc/}.

\clearpage
\bibliographystyle{plainnat}
\bibliography{refs}

\end{document}